\documentclass[conference]{IEEEtran}
\IEEEoverridecommandlockouts
\usepackage{cite}
\pdfoutput=1
\usepackage{amsmath,amssymb,amsfonts}
\usepackage{algorithmic}
\usepackage{graphicx}
\usepackage{textcomp}
\usepackage{xcolor}
\def\BibTeX{{\rm B\kern-.05em{\sc i\kern-.025em b}\kern-.08em
    T\kern-.1667em\lower.7ex\hbox{E}\kern-.125emX}}
\newcommand{\RN}[1]{%
  \textup{\uppercase\expandafter{\romannumeral#1}}%
}
\begin{document}

\title{FairMixRep : Self-supervised Robust Representation Learning for Heterogeneous Data with Fairness constraints
}



\author{\IEEEauthorblockN{Souradip Chakraborty}
\IEEEauthorblockA{Walmart Labs\\
Souradip.Chakraborty\\@walmartlabs.com
}
\and
\IEEEauthorblockN{Ekansh Verma}
\IEEEauthorblockA{Walmart Labs\\
Ekansh.Verma\\@walmartlabs.com}
\and
\IEEEauthorblockN{Saswata Sahoo}
\IEEEauthorblockA{Gartner\\
saswata.sahoo\\@gartner.com}
\and
\IEEEauthorblockN{Jyotishka Datta}
\IEEEauthorblockA{University of Arkansas\\
jd033@uark.edu}
}

\maketitle

\begin{abstract}
 Representation Learning in a heterogeneous space with mixed variables of numerical and categorical types has interesting challenges due to its complex feature manifold. Moreover, feature learning in an unsupervised setup, without class labels and a suitable learning loss function, adds to the problem complexity. Further,  the learned representation and subsequent predictions should not reflect discriminatory behavior towards certain sensitive groups or attributes. The proposed feature map should preserve maximum variations present in the data and needs to be fair with respect to  the sensitive variables. 
 We propose, in the first phase of our work, an efficient encoder-decoder framework to capture the mixed-domain information.  The second phase of our work focuses on de-biasing the mixed space representations by adding relevant fairness constraints. This ensures minimal information loss between the representations before and after the fairness-preserving projections. Both the information content and the fairness aspect of the final representation learned has been validated through several metrics where it shows excellent performance. Our work (FairMixRep) addresses the problem of Mixed Space Fair Representation learning from an unsupervised perspective and learns a Universal representation which is timely, unique and a novel research contribution. 
 
 \footnote{This paper is accepted at ICDM'2020 DLC Workshop.}. 
 
\end{abstract}


\section{Introduction}

With the widespread use of artificial intelligence and machine learning (AI/ML) in our everyday life, the need to ensure fairness in these systems has come to the fore, with an overarching goal that the outputs and decisions do not reflect discriminatory behavior toward certain sensitive groups or attributes. The purpose is to learn representations which are independent of sensitive attributes including but not limited to race, gender, ethnicity etc. Despite the rapid and influential growth in the literature on representation learning, essentially no work has been done to ensure group fairness in the context of a Heterogeneous mixed space representation learning problem, and it remains a both an unexplored and timely area of research at this crucial juncture of evolution of AI/ML literature. 
\par
Heterogeneous mixed type of variables consisting of numerical and categorical variables assume values in a complex manifold.  Numerical variables take values on the real line within the range of variations whereas categorical variables indicate class label information. Multi-level categorical variables can be assumed to be binary indicator variables taking values $1$ or $0$, depending on presence or absence of each of the levels in a data point. For mixed space, prior works on feature learning, especially in an unsupervised setup is relatively unexplored, which explicitly incorporates the marginal mixed space information and mutual dependence structure.
\par
It is also important to ensure that derived feature maps are uncorrelated with sensitive subject specific variables to guarantee fairness in learning. The learned features and subsequent outcomes should not be biased towards subject traits which might spuriously influence the learning outcome. For example, feature map on salary data and related mixed variables should not be influenced by subject traits such as gender, ethnicity, race or religion. Such fair representations of the mixed variables in the derived feature map with respect to sensitive variables have not been explored before in similar setup. 
\par
There are multiple possibilities from which the concept of fairness can be addressed. In this work, we address the group fairness issue based on a guided projection of latent mixed space feature embedding derived from an encoder-decoder architecture. We ensure that the new fair representations will be independent of the sensitive variables with minimum information loss. In the first phase of our work, we propose a novel nonlinear Encoder-Decoder framework to capture the cross-domain information for mixed data types in a latent feature embedding. The hidden layers of the network  connect  each of the two types of variables through a sequence of non-linear  transformations to give the latent feature representation. There are two separate networks switching the roles of the numerical and categorical variables as network input and output to get the latent feature embedding. We quantitatively evaluate the quality of our embeddings learned by a simple linear evaluation supervised model. In the second phase of our work, we study the bias and fairness aspects in our learned representations. We de-bias our mixed space representations learned through a fair projection with relevant fairness constraints. 
\par
Our work (FairMixRep) addresses the problem of Mixed Space Fair Representation learning from an unsupervised perspective and learns a Universal representation which is a novel research work and shows to perform exceptionally well in the Adult income and German credit data sets.
\par
\noindent \textbf{Contributions}: The main contributions of our work are :
\begin{enumerate}
  \item We create a modular framework to generate robust representation of the mixed space data with continuous and categorical variables and ensure its fairness with guided projections without using any label information.
  \item We ensure the group fairness aspect in our representations learned for sensitive groups like gender, ethnicity, race etc. Our methodology is capable of producing unbiased representation with multiple sensitive attributes by minimizing their effect simultaneously.
  \item We validate our learned representations FairMixRep with a simple logistic regression classifier on two real world datasets. Our proposed methodology achieves almost similar accuracy with ensured fairness.
  \item FairMixRep gives a universal representation for the mixed space data without using any label information and can be further easily fine-tuned for downstream tasks such as classification as shown in our work.
\end{enumerate}

\par
\noindent \textbf{Use Cases}: Our methodology provides a natural framework to ensure group-fairness for the learned representations for sensitive covariates like gender, ethnicity, race etc. The purpose of ensuring group fairness is to mitigate the disparity that happens in critical decision making for individuals belonging to such sensitive groups. The decisions can include giving loans, granting admission to students, giving imprisonment etc. which are very critical and even a small amount of bias might adversely affect an individual belonging to historically marginalized groups \cite{suresh2019framework}. Majority of real world datasets have mixed attributes \textit{i.e.} both categorical and continuous variables and the data might not be labelled or can have a small subset of labeled data points. In such cases, the methodology becomes heavily dependent on the representations learned. For example : when a bank opens up its branch at a new location and if the loan approval system is dependent on a machine learning model, then it is expected that initial level of predictions will be inaccurate and biased. Here, the demographic attributes will have both categorical and numerical variables as well as sensitive information. In such cases, FairMixRep provides an accurate starting point by creating an unbiased and robust representation which can be later fine-tuned for various downstream tasks (e.g. predicting the appropriate loan rating) based on available data.

\section{Previous Work}
There have been extensive research in the field of Unsupervised representation learning ranging over global methods such as linear projection on the  principal component space \cite{pca,spca}, non-linear feature map on Kernel space (see \cite{kpca}) or locality based manifold learning
such as locally linear embedding \cite{lle}, isometric feature mapping \cite{isomap} etc.  Multidimensional scaling \cite{de1980multidimensional} gives feature representation \cite{mds} preserving mutual dissimilarity. Nonlinear feature maps can be derived using various nonlinear functions in iterative fashion to incorporate different degrees of non-linearity in a deep learning framework(see ~\cite{dl}). 
\par
For mixed data types in unsupervised learning, discretization  of numerical variables and treating all the variables as categorical type is proposed by  \cite{discretize}. Similarly categorical variables can be converted to numerical type by utilising the intrinsic low rank structure and dense embedding (see e.g. \cite{catembed1},\cite{catembed2}, \cite{catembed3}). Considering different distance metrics for mixed variables was proposed by \cite{kononenko}. Representations of mixed variables as nodes of a weighted undirected graph followed by Laplacian embedding was proposed by \cite{sahoo}. Learning latent representations for mixed data types with a nonlinear Deep Encoder-Decoder Framework followed by a locality preserving projection was proposed by \cite{sahoo2020learning}.
\par
Learning representations that are useful for predictions in various downstream tasks and additionally not be discriminatory against sensitive attributes is thus an important but a challenging methodological and practical problem. Solutions proposed to learning fair and invariant representations have a long history. Several approaches which leverage  the  Generative  Adversarial  Networks  (GANs) \cite{NIPS2014_5423} have been proposed to learn robust fair and transferable representations (e.g. \cite{9aa5ba8a091248d597ff7cf0173da151},\cite{DBLP:journals/corr/abs-1802-06309},\cite{xie17nips},\cite{DBLP:journals/corr/abs-1801-07593}). General methodology formulated in these approaches is to optimize the encoder that learns the meaningful representation and an adversary which extracts sensitive attributes from the learned representation. Together they are involved in a min-max game, solution to which correspond to fair representations. \cite{DBLP:journals/corr/abs-1805-09458} show that adversarial training is unnecessary and sometimes counter-productive. Further they derived a variational upper bound for the mutual information between latent representations and sensitive attributes. \cite{DBLP:journals/corr/abs-1906-02589} have proposed a fair representation learning method  by  \textit{disentanglement},  which  can  be  modified  at  test  time  to yield  a  fair representation with respect to multiple sensitive attributes and their conjunctions, even when test-time sensitive attribute labels are unavailable. Recently, \cite{zhao2019conditional} proposed  a  representation learning algorithm that aims to  simultaneously ensure accuracy parity and equalized odds. The main idea underlying their algorithm is to align the conditional distributions of representations and use balanced error rate on both the target variable and the sensitive attribute. 

\section{Proposed Methodology}
In this section we describe the architecture of FairMixRep : Self-supervised Robust Representation Learning for Heterogeneous Data with Fairness constraints in details. In the first stage of our work we learn the mixed space robust representation of the feature space and in the second stage we ensure fairness in our mixed space representation through relevant fairness constraints and guided projections. To comply with disparate treatment criterion, we are not using the sensitive attributes while learning the mixed space representations in the encoder-decoder framework.

\subsection{Self-Supervised Heterogeneous Representation Learning} Consider Fig. 1 for the network architecture of stage 1 of the proposed methodology. Motivated by \cite{sahoo2020learning}, we create an encoder-decoder network which takes the numerical data points as input and forms an encoded latent representation. Subsequent non-linear projections helps in reconstruction of the categorical variable in the decoding phase of this network. We create a similar network, where the input is now the categorical variables and the latent encoded representation helps in reconstruction of the continuous variable in the decoding phase of this network. The latent encoder feature representations from both the networks are concatenated to give our final mixed space representation. 

\par
Denote $i^{th}$ row of the numerical variable data matrix $\mathcal{D}^{(num)} \in \mathbb{R}^{n\times d_{1}}$  by $x^{(num)}_{(i)}$ and that of the categorical variable data matrix $\mathcal{D}^{(cat)} \in \mathbb{R}^{n \times d_{2}}$ by $x^{(cat)}_{(i)}$. Here $d_{1}$ \& $d_{2}$ represent the dimensions of the numerical data space and the categorical data space respectively and $n$ represents the total number of observations. In our work, we have encoded the multi-level categorical variables as dummy binary indicator variables taking values $1$ or $0$, depending on presence or absence of each of the levels in an observation. 
\par
We learn two encoder-decoder based networks $N_{num-cat}$ \& $N_{cat-num}$ that helps in efficiently generating a latent space representation of our heterogeneous feature space. The numerical variable data point $x^{(num)}_{(i)}$ is taken as an input to the network $N_{num-cat}$. It is then mapped to a latent encoder representation $z^{(num)}_{(i)} = f(x^{(num)}_{(i)})$, where $m$ is the intermediate latent encoder dimension for $N_{num-cat}$. 
The latent encoder representation $z^{(num)}_{(i)}$ of dimension $d_{\kappa_1}$ now helps in reconstruction of the $i^{th}$ observation on categorical variable $x^{(cat)}_{(i)}$ in the decoding phase of this network. The reconstruction loss for the network $N_{num-cat}$, represented by $L_{num-cat}$ is a binary cross-entropy loss function averaged over each dimension of the categorical variable. The average reconstruction loss $L_{num-cat}$ is minimized to learn the parameters for this network. 

\par
Similarly, for the $N_{cat-num}$ network, the categorical variable data point $x^{(cat)}_{(i)}$ is taken as an input where it is mapped to a latent encoder representation $z^{(cat)}_{(i)} = g(x^{(cat)}_{(i)})$, $n$ being the intermediate latent encoder dimension for $N_{cat-num}$. The latent encoder representation $z^{(cat)}_{(i)}$ of dimension $d_{\kappa_2}$, now helps in reconstruction of the $i^{th}$ numerical variable data point $x^{(num)}_{(i)}$ in the decoding phase of this network. The reconstruction loss for the network $N_{cat-num}$, represented by $L_{cat-num}$ is a mean squared error loss function over each dimension of the numerical variable. The average reconstruction loss $L_{num-cat}$ is minimized to learn the parameters for this network. 
\par
The latent encoder feature representations $z^{(num)}_{(i)}$, $z^{(cat)}_{i}$ from the networks $N_{cat-num}$, $N_{num-cat}$ respectively are concatenated to give our final mixed space representation $z^{(conc)}_{(i)} = [z^{(num)}_{(i)},z^{(cat)}_{(i)}] \in R^{p}$,  where $p =  d_{\kappa_1}+d_{\kappa_2}$.

\begin{figure}[hbt!]
\centering
\includegraphics[width=9cm]{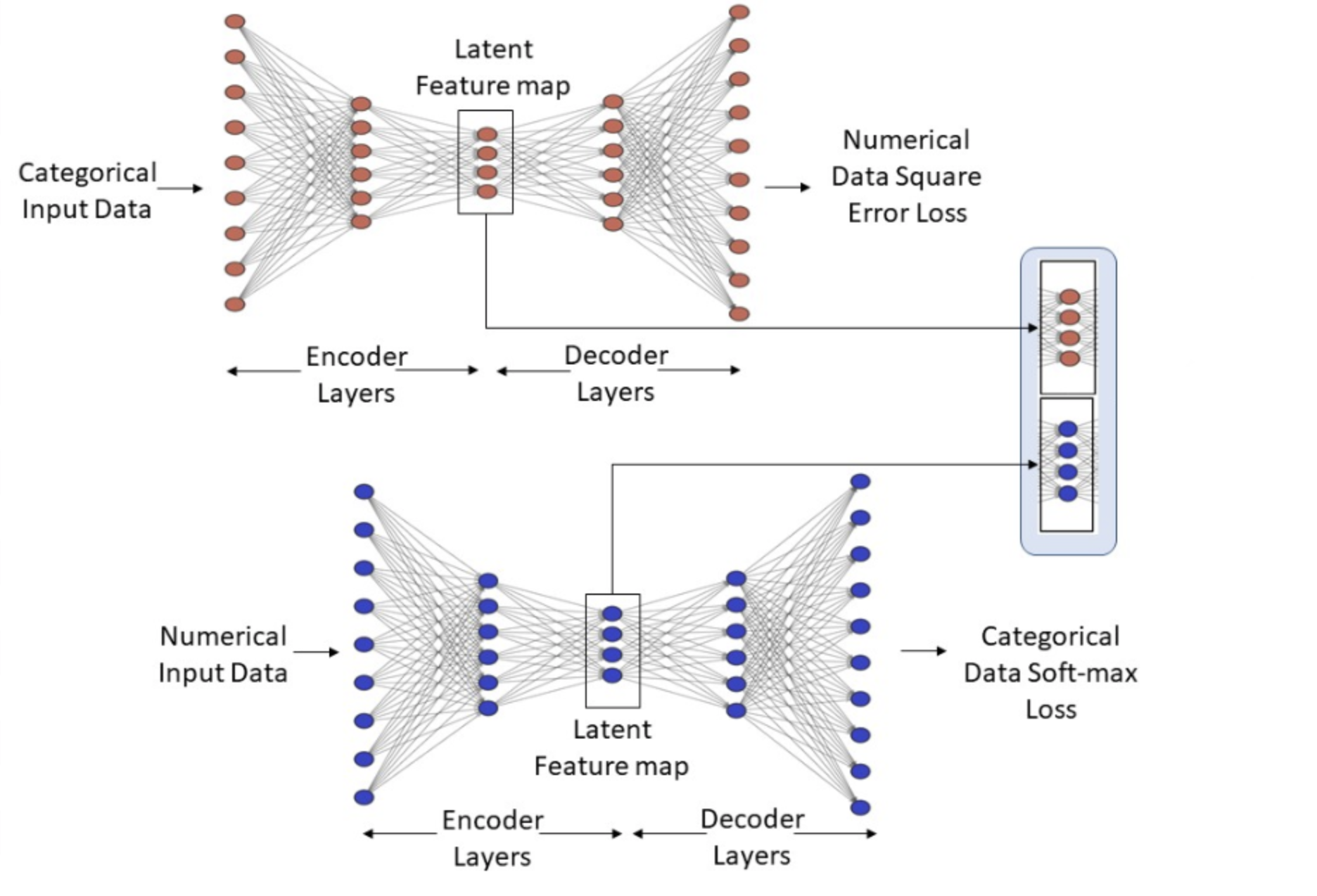}
\caption{ Mixed Space Encoder-Decoder Network}
\label{fig:architecure}
\end{figure}

\subsection{Fairness Constraints in Representation Learning} In the stage 2 of our work, we study the bias  in our learned mixed space representation $z_{conc}$ from a fairness perspective. The purpose is to learn representations which are independent of sensitive attributes like race, gender, ethnicity etc.
\par
Let us denote $Z=[z^{(conc)'}_{(i)}, i=1,2, \ldots, n] \in \mathbb{R}^{n\times p}$ as the current representations of our input data matrix. Here $n$ represents the number of data points and $p$ represents the dimension of the mixed space representation, where $p$ = $d_{\kappa_1}+d_{\kappa_2}$. Let $S$ be the sensitive group membership feature space with sensitive attributes. Our objective in this phase is to ensure fairness in our representation $Z$ with respect to the sensitive feature space $S$. We try to project the $Z$ matrix in a space which is orthogonal to the feature space of $S$ with minimal information loss. In solving the above problem, we take reference from the work by \cite{aliverti2018removing} on removing the influence of a group variable in high-dimensional
setting.
\par
In the current scenario, we try to estimate $\hat{Z} \in \mathbb{R}^{n\times p}$ which is a reconstruction version of the mixed space representation $Z \in \mathbb{R}^{n\times p}$ such that  $\hat{Z}$ is orthogonal to the space of $S$ with minimal information loss. It has been often observed that in high-dimensional problems, there is a low-rank representation of the feature space which significantly captures the maximum information. Hence, as shown in \cite{aliverti2018removing}, we express the reconstruction representation as $\hat{Z}$ = $UV^T$ where $V \in \mathbb{R}^{p\times k}$, consisting of $k$ orthonormal basis vectors and $U \in \mathbb{R}^{n\times k}$ gives the association scores. The problem of learning fair representation $\hat{Z}$ gets reduced to solving the problem of minimizing the reconstruction error under fairness constraints: $\left\Vert Z - U V^T\right\Vert^2_{F}$ subject to $<\hat{Z},S> = 0$. Here $\hat{Z} = U V^T$ and $V \in G_{p,k}$. $G_{p,k}$ is the Grassmann manifold of orthonormal matrices. It has been shown in \cite{aliverti2018removing} that this problem can be exactly solved and has a closed form solution. Let, the rank-k singular value decomposition of the $Z$ matrix is given by $M_{k}D_{k}N_{k}^{T}$. Then the exact solution for the representations satisfying the condition of orthogonality to sensitive groups with minimal information loss is given by $\hat{Z} = (I_{n} - P_{S})M_{k}D_{k}N_{k}^{T}$, where $P_{S} = S(S^TS)^{-1}S^T$. $\hat{Z}$ is the final fair latent feature map for our heterogeneous mixed space representation learning problem. We perform a detailed quantitative evaluation of our representations learned both from an information content perspective and fairness perspective in the next section. We show that the final mixed space fair representations learned (FairMixRep) performed exceedingly well with respect to both the aspects.

\section{Datasets}
We evaluate our proposed methodology on two real world datasets which are provided by UCI ML-repository \cite{Dua:2019}.
The first, the Adult income dataset has $n = 45,222$ data points and the objective is to predict whether a person has savings of over $50,000$ USD with the sensitive attributes being corresponding race and gender. We use both categorical and continuous columns as provided in the dataset. 
The second, the German credit dataset has $n = 1,000$ instances which classify bank account holders into credit class good or bad indicated by the probability of default for an individual. Each person is described by $20$ attributes.  In our experiments we consider age as the sensitive attribute by discretizing it into two groups: 25 $\leq$ age $\geq$ 60; age $<$ 25 and age $>$ 60.
Details on the input features to the proposed methodology are provided in Table \ref{tab:data}.
\begin{table}[hbt!]
\centering
\caption{Dataset}
\label{tab:data}
\begin{tabular}{|c|c|c|c|} 
\hline
\multicolumn{1}{|c|}{Dataset} & \multicolumn{1}{|c|}{num-Cont} & \multicolumn{1}{|c|}{num-Cat} & \multicolumn{1}{|c|}{num-Sensitive}   \\
\hline
 Adult &6 & 6 & 2: \textit{race}, \textit{sex} \\
\hline
 German  & 6 & 13 & 1: \textit{age} \\
\hline
\end{tabular}
\end{table}

\section{Experimental Setup}
Our experimental setup is broadly divided into two major parts, Mixed Space Representation Learning and Unsupervised Representation Learning with fairness constraints.

\textbf{\textit{Mixed Space Representation Learning}} :
We vary the choices of the $k_1$, $k_2$ and $p$ to obtain the feature maps for different datasets. To evaluate the  discriminatory power of feature maps we project them to produce class logits by adding a linear classification head on top. We keep the features fixed which implies that only linear classification head has trainable weights. We create a stratified split based on class label and use 50\% of the data for testing purposes. 
Network hyper-parameters for both, German credit and Adult income are detailed in table \ref{tab:params} 
We used binary cross-entropy for $N_{num-cat}$ and mean squared error for $N_{cat-num}$ as loss functions to train the respective networks.
We used Adam optimizer with learning rate value of 0.001 for both $N_{num-cat}$ and $N_{cat-num}$ across the datasets.
\begin{table}[hbt!]
\centering\caption{ NUMBER OF HIDDEN LAYERS ($k_1$, $k_2$) AND MIXED SPACE REPRESENTATION DIMENSION ($p$)}
\label{tab:params}
\begin{tabular}{|c|c|c|c|} 
\hline
\multicolumn{1}{|c|}{Dataset} & \multicolumn{1}{|c|}{$k_1$} & \multicolumn{1}{|c|}{$k_2$} & \multicolumn{1}{|c|}{$p$}   \\
\hline
Adult Dataset & 5 & 5 & 200  \\
\hline
German Credit Dataset  & 3 & 3 & 100 \\
\hline
\end{tabular}
\end{table}

\begin{figure}[hbt!]
\includegraphics[width=9cm]{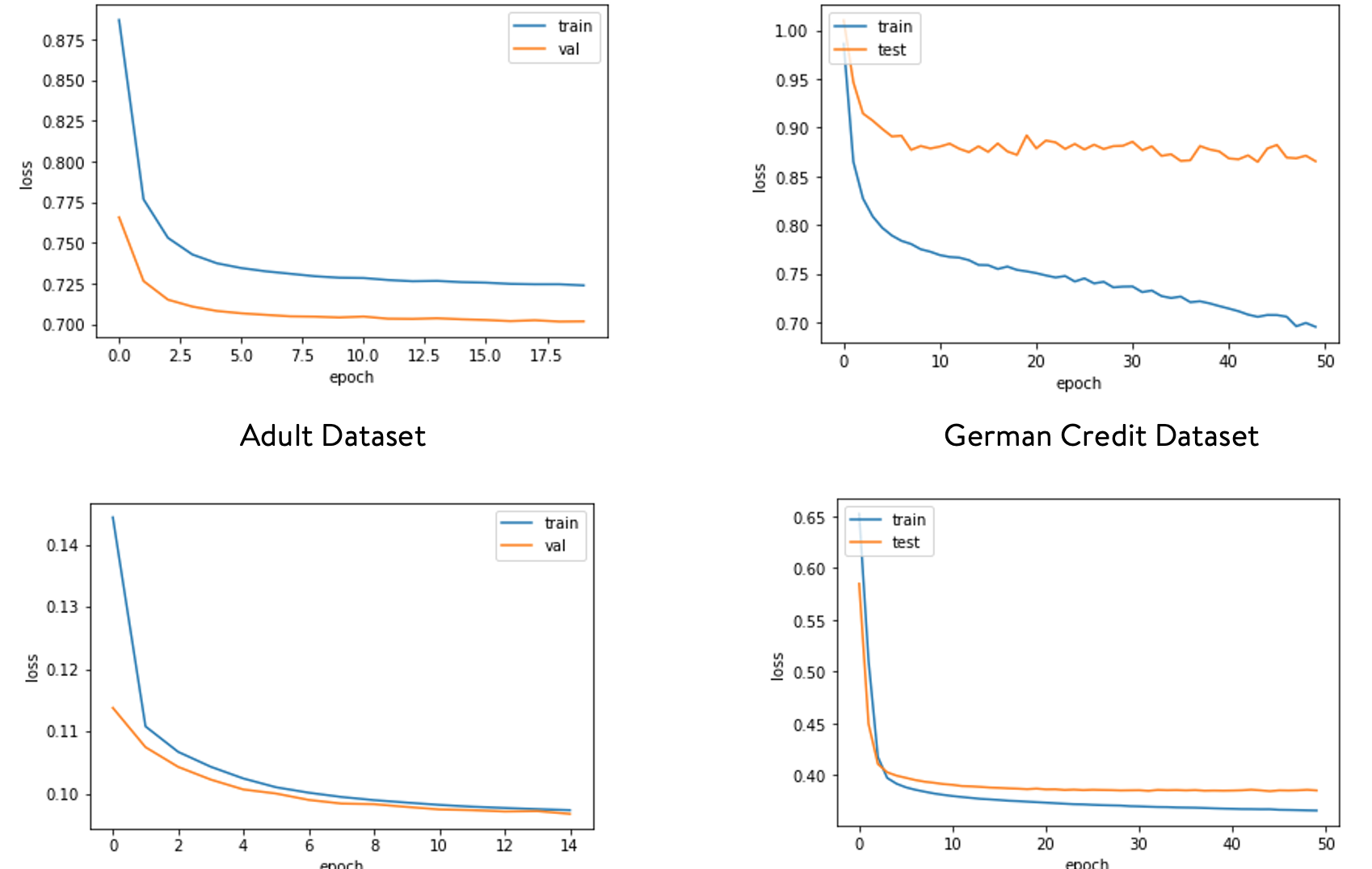}
\caption{ The values of the soft-max binary cross entropy loss and means square error loss function for the different epochs of estimating the deep hidden layer networks are give for the 4 datasets. The plots at the top indicate mean square error loss for the network where the input is categorical data and output is the numerical data. The plots at the bottom indicate binary cross entropy loss when the input is numerical data and output is the categorical data. The blue curve denotes the training loss and the orange curve denotes the validation loss.}
\label{fig:loss-function}
\end{figure}

\textbf{\textit{Representation Learning with Fairness constraints}} :  
We performed rank-k singular value decomposition of the learned mixed space representations as first step in the process of removing the effect of sensitive attributes. Values of $k$ chosen for Adult and German dataset are 18 and 20 respectively. Possible values of $k$ are selected using the percent variability explained by the singular vectors for a given dataset.

\section{Results \& Numerical Investigation}
In this section, we do a detailed study to quantitatively evaluate the quality and performance of \textit{FairMixRep : Self-supervised Robust Representation Learning for Heterogeneous Data with Fairness constraints}.
The quantitative evaluation and investigation of our learned representations also happen in two phases. In the first phase, we primary evaluate the quality of the latent representation from an information content perspective and in the second phase from a Fairness and Bias perspective.

\subsection{Evaluation Metrics}
Firstly, we evaluate the quality of the latent mixed space representations learned in an unsupervised setting. In the unsupervised mode of evaluation, we mainly observe the convergence of the self-supervised loss for the Encoder-Decoder network. For the network $N_{num-cat}$, the self-supervised loss is an average binary cross-entropy loss whereas for the network $N_{cat-num}$, the self-supervised loss is mean squared error loss. 

\begin{figure*}[hbt!]
\includegraphics[height=9cm]{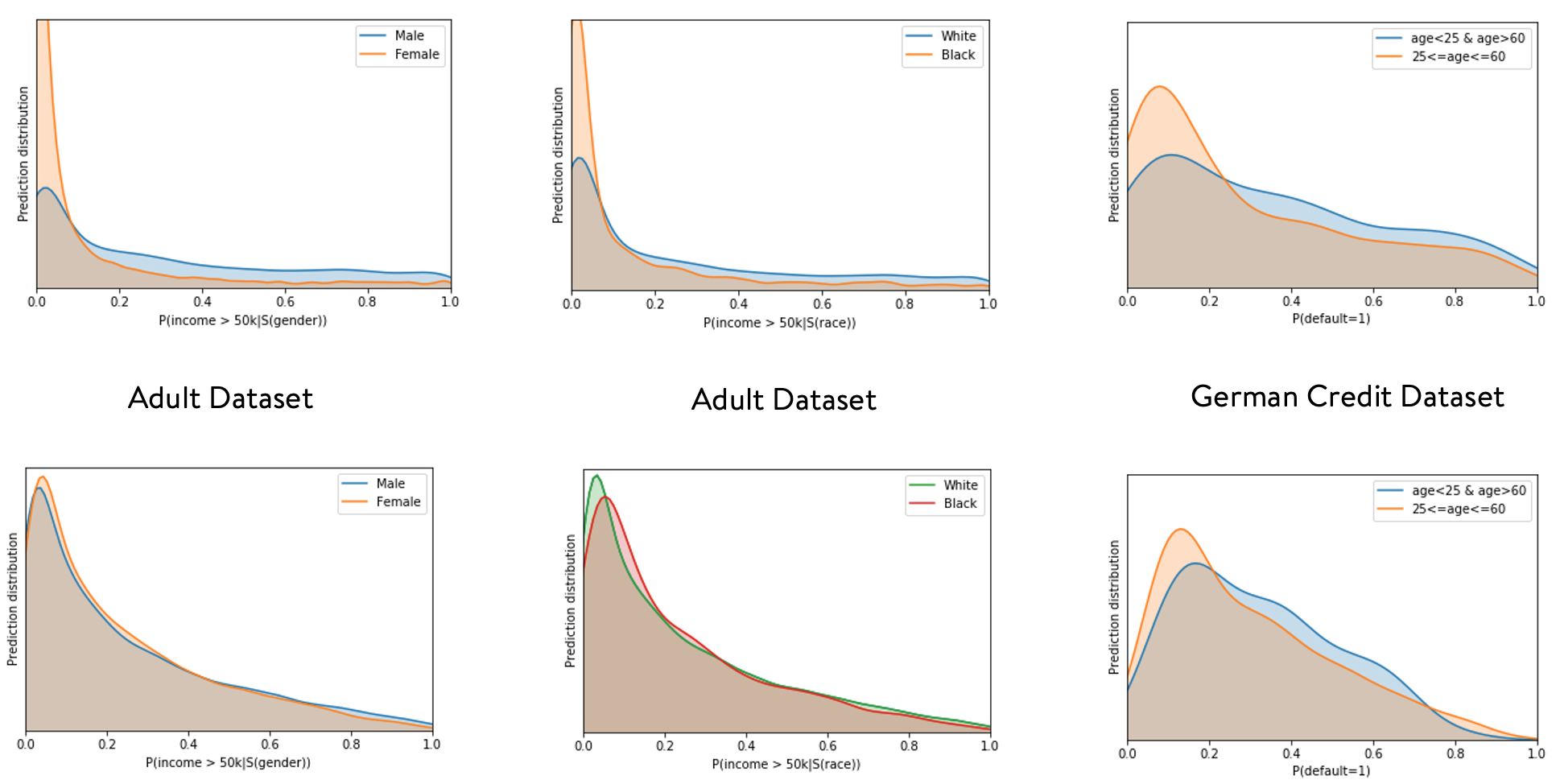}
\caption{ Prediction distribution of the label space through a Logistic Regression model.The plots at the top indicate the high bias in the learned mixed space representation. The plots at bottom indicate the fair representations after using fairness constraints on the learned representations.}
\label{fig:loss-p-rule}
\end{figure*}

To quantitatively evaluate our learned representations, we add a logistic regression model on top of the learned embeddings to predict the true labels. Further we calculate the validation accuracy and ROC-AUC score for the binary logistic regression and report results in the table. We understand that having a more complex supervised model and fine-tuning the solution will definitely enhance the accuracy performance on the true labels space. But the objective in this case is to evaluate the quality of the unsupervised mixed space representation learned and hence we have used simple logistic regression on top of the learned mixed space representation to evaluate the performance measure.

\subsection{Fairness Metrics}

Various quantitative fairness metrics have been introduced \cite{romei_ruggieri_2014}, \cite{DBLP:journals/corr/abs-1712-03586} based on different definitions of fairness.

\begin{itemize}
    \item \textbf{Disparate Impact (DI)} Ratio of probability of unprivileged group getting positive prediction to the probability of privileged group getting positive prediction.
    \[DI = \frac{P(\hat{Y} = 1\mid S = 0)}{P(\hat{Y} = 1\mid S = 1)}\]
    
    \item \textbf{Statistical Parity Difference (SPD)} Instead of the ratio of probabilities as shown in DI, the difference is calculated.
    \[SPD = P(\hat{Y} = 1\mid S = 0) - P(\hat{Y} = 1\mid S = 1)\]
    
    
    
    
\end{itemize}

In our evaluation for disparate impact we 
adopt the 80\% rule \cite{biddle2005adverse}, \cite{zafar2015fairness} which states that a model is fair when the value of $DI\times100$ is greater than 80\%.


\subsection{Quantitative Evaluation and Performance of the Mixed Representation}
We report the performance of mixed representation in predicting the true class label by utilising the output from logistic regression model trained on top of the representation. The prediction power of learned representations using the proposed self supervised methodology can be noticed from its near equivalent performance as compared to the supervised counterparts \cite{zafar2017parity},\cite{DBLP:journals/corr/abs-1801-07593},\cite{jiang2019wasserstein} on the binary classification across both the datasets. The near-equivalent performance is achieved in spite of training the logistic classifier with only 50\% of the total labeled data available.

\subsection{Evaluation of the Fairness aspects in the Representation learned}
Fairness metrics discussed earlier are calculated for the representations before and after taking advantage of the fairness constraints.
Our experimental findings are displayed in Table \ref{tab:bias} and Table \ref{tab:fair}.
We observe that de-biasing has a small detrimental effect on overall prediction power of mixed space representation (accuracy : $0.85$ to $0.81$ for Adult income prediction). De-biased mixed space representation under fairness constraints obey Disparate Impact and Statistical  Parity  Difference for the sensitive attributes across both, Adult income and German credit datasets.

\begin{table}[hbt!]
\centering\caption{ Fairness metrics for biased representations}
\label{tab:bias}
\begin{tabular}{|c|c|c|c|c|c|} 
\hline
\multicolumn{1}{|c|}{Dataset} &
\multicolumn{1}{|c|}{Accuracy} &
\multicolumn{1}{|c|}{ROC-AUC} &
\multicolumn{1}{|c|}{Sensitive} & \multicolumn{1}{c|}{DI*100} &
\multicolumn{1}{|c|}{SPD}  \\

\hline
Adult & 0.85 & 0.91 & Gender & 33.17 & 0.172 \\
\hline
Adult & 0.85 & 0.91 & Race & 42.50 & 0.122   \\
\hline
German Credit & 0.718 & 0.74 & Age & 70.52 & 0.089 \\

\hline
\end{tabular}
\end{table}

\begin{table}[hbt!]
\centering\caption{ Fairness metrics for de-biased representations}
\label{tab:fair}
\begin{tabular}{|c|c|c|c|c|c|} 
\hline
\multicolumn{1}{|c|}{Dataset} &
\multicolumn{1}{|c|}{Accuracy} &
\multicolumn{1}{|c|}{ROC-AUC} &
\multicolumn{1}{|c|}{Sensitive} & \multicolumn{1}{|c|}{DI*100} & 
\multicolumn{1}{|c|}{SPD}  \\
\hline
Adult &0.81 & 0.84 & Gender & 85.34 & 0.026 \\
\hline
Adult &0.81 & 0.84 & Race & 87.29 & 0.021   \\
\hline
German Credit & 0.734 & 0.74 & Age & 90.90 & 0.0314 \\

\hline
\end{tabular}
\end{table}


\par

\section{Conclusion}
Our work, FairMixRep, addresses the problem of Mixed Space Fair Representation learning from an unsupervised perspective and learns a universal representation which is a novel field of research. We create a simple yet powerful framework to generate robust representation of the mixed space data and ensure its fairness with guided projections without using any label information. Moreover, our methodology is capable of producing the unbiased representation with multiple sensitive attributes by minimizing their effect simultaneously. Finally, we validate our learned representations with a simple logistic regression classifier on two real world datasets and our proposed framework achieves almost similar accuracy before and after the fair projections. 

For future work, we would like to extend and evaluate FairMixRep for other critical use cases such as fair clustering and fair anomaly detection. A particularly promising and important application area is bias in predictive policing \cite{perry2013predictive, ferguson2016policing, lum2016predict} and it will be worthwhile to apply FairMixRep in that context. 


\bibliographystyle{IEEEtran}
\bibliography{biblio_mixed}
\end{document}